\def\year{2022}\relax
\documentclass[letterpaper]{article} % DO NOT CHANGE THIS
\usepackage{aaai22}  % DO NOT CHANGE THIS
\usepackage{times}  % DO NOT CHANGE THIS
\usepackage{helvet}  % DO NOT CHANGE THIS
\usepackage{courier}  % DO NOT CHANGE THIS
\usepackage[hyphens]{url}  % DO NOT CHANGE THIS
\usepackage{graphicx} % DO NOT CHANGE THIS
\usepackage{natbib}  % DO NOT CHANGE THIS AND DO NOT ADD ANY OPTIONS TO IT
\usepackage{caption} % DO NOT CHANGE THIS AND DO NOT ADD ANY OPTIONS TO IT
\DeclareCaptionStyle{ruled}{labelfont=normalfont,labelsep=colon,strut=off} % DO NOT CHANGE THIS
\usepackage{algorithm}
\usepackage{algorithmic}

\usepackage{amsmath}
\usepackage{latexsym}
\usepackage{booktabs}
\usepackage{url}
\usepackage{amsmath}

\usepackage{graphicx}

\usepackage{microtype}

\usepackage{graphicx}
\usepackage{amsmath}
\usepackage{amsfonts}
\usepackage{array}
\usepackage{comment}
\usepackage{url}
\usepackage{amsmath}
\usepackage{mathtools}

\usepackage{graphicx}
\usepackage{amsmath}
\usepackage{amsfonts}
\usepackage{array}
\usepackage{comment}

\usepackage{newfloat}
\usepackage{listings}
\floatstyle{ruled}
\newfloat{listing}{tb}{lst}{}
\floatname{listing}{Listing}
\title{Improving Keyphrase Extraction\\with Data Augmentation and Information Filtering}
\author{
    Amir Pouran Ben Veyseh, \textsuperscript{\rm 1}\\
    Nicole Meister, \textsuperscript{\rm 2} Franck Dernoncourt, \textsuperscript{\rm 3} and Thien Huu Nguyen\textsuperscript{\rm 1}
}
\affiliations{
    \textsuperscript{\rm 1}Department of Computer and Information Science, University of Oregon\\

    \textsuperscript{\rm 2}Department of Electrical and Computer Engineering, Princeton University\\

    \textsuperscript{\rm 3}Adobe Research\\

    apouranb@cs.uoregon.edu
}

\usepackage{bibentry}
\begin{document}

\maketitle

\begin{abstract}
Keyphrase extraction is one of the essential tasks for document understanding in NLP. While the majority of the prior works are dedicated to the formal setting, e.g., books, news or web-blogs, informal texts such as video transcripts are less explored. To address this limitation, in this work we present a novel corpus and method for keyphrase extraction from the transcripts of the videos streamed on the Behance platform. More specifically, in this work, a novel data augmentation is proposed to enrich the model with the background knowledge about the keyphrase extraction task from other domains. Extensive experiments on the proposed dataset dataset show the effectiveness of the introduced method.
\end{abstract}

\section{Introduction}

Keyphrases are one or different continuous words that may speak to the most thoughts in a report. Keyphrases are commonly categorized as Present or Absent. A present keyphrase unequivocally shows up within the document, while an Absent keyphrase does not exist within the record. Keyphrases can serve as brief rundown for a archive, thus profiting different NLP applications Information Recover \cite{hersh2021information} and Content Summarization \cite{adhikari2020exploring}. Due to their value, within the more than two decades, KP has been considered in numerous inquire about works \cite{turney2000learning,wu2005domain,jiang2009ranking,hasan2014automatic,mahata2018key2vec,chen2020exclusive,ye2021one2set}.

Though customarily highlight designing has been utilized for KP \cite{turney2000learning,sheeba2014fuzzy}, as of late profound learning is demonstrated to be more productive for this errand \cite{ye2021one2set,ahmad2021select}. Be that as it may, one restriction within the current works is that they are basically restricted to the formal content such as logical papers \cite{meng2017deep} and web-logs \cite{xiong2019open}. As such, the challenges in other spaces are still uncertain. Among others, video transcript is one of the less-explored spaces that may altogether advantage from KP. For occurrence, it may be utilized for video summarization and recovery or advantage individuals who are hard of hearing and difficult of hearing (DHH) \cite{kafle2019evaluating}. On the other hand, KP for transcripts that are consequently gotten are more challenging than the formal composed records as these transcripts include loud content, incomplete/repeated sentences and expressions, casual lexicon, and non cohesive data stream. In spite of the fact that there have been many related endeavors to assess highlight designing strategies on assembly transcripts \cite{sheeba2014fuzzy,sheeba2012improved}, the accessible assets, with a modest bunch of transcripts and keyphrases, are not valuable to train/evaluate the later progressed profound models.

\begin{figure*}
    \centering
    \includegraphics[scale=0.4]{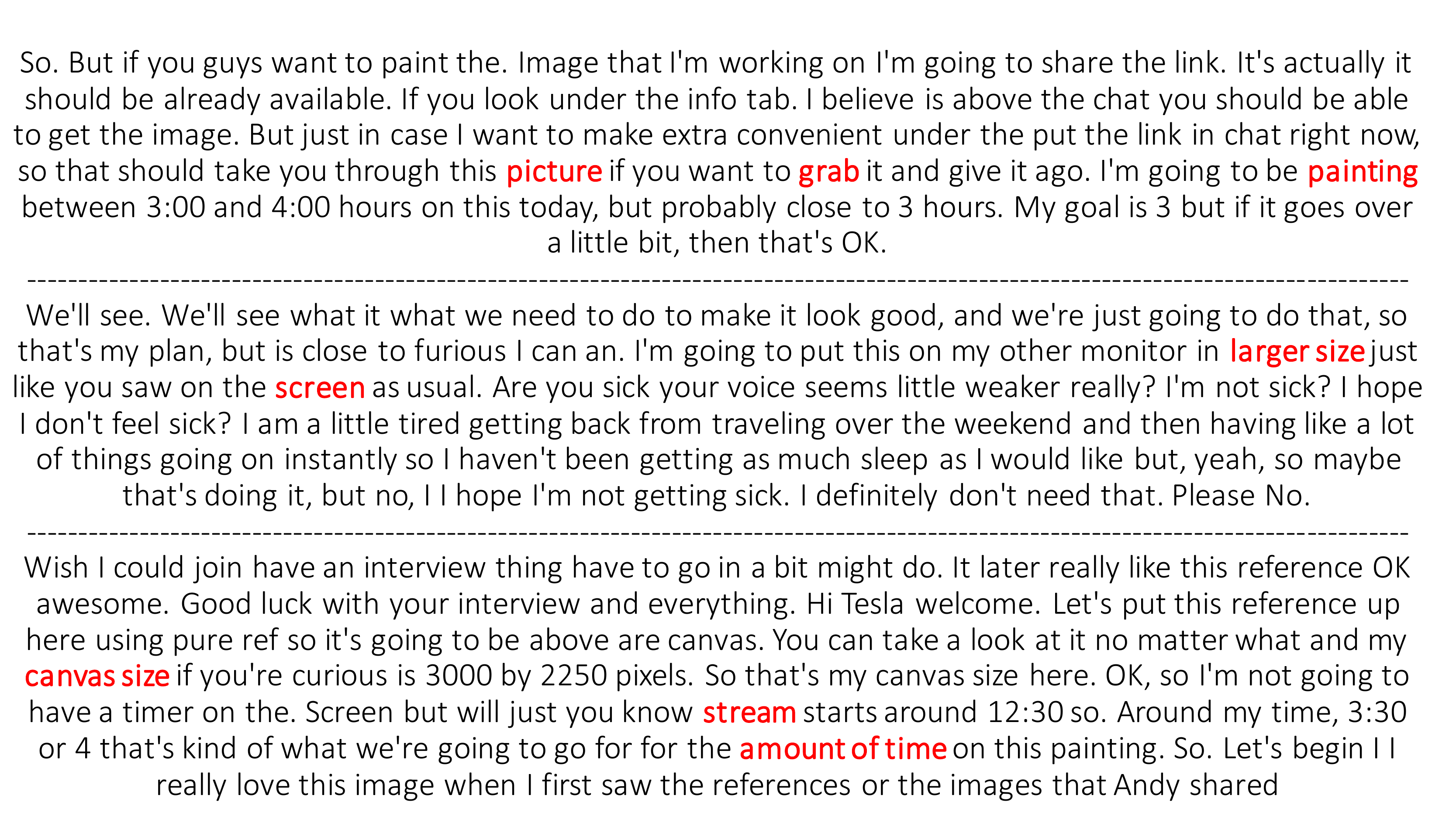}
    \caption{Part of the transcript of a live-streamed video. The paragraphs are separated by dashed lines and the keyphrases are shown in red boldface.}
    \label{fig:example}
\end{figure*}

To address these limitations, in this work, we study the task of keyphrase extraction in the domain of video transcripts. In particular, we explore the transcript of the videos streamed on the Behance\footnote{www.behance.net}. An example of these transcripts are presented in Figure \ref{fig:example}. To this end, first a collection of the video transcripts are annotated with keyphrases. Next, we explore deep learning models to extract the keyphrases from the the transcripts of the videos. One of the challenges for the keyphrase extraction in video transcripts is that the amount of training samples in this domain is less than the well-studied domains such as news. As such, in a novel approach, in this work we present a data augmentation technique, in which the data from other domains is employed to improve the performance of the keyphrase extraction model. Our extensive analysis on the proposed dataset reveal the effectiveness of the proposed approach.

\section{Related work}

This task could be modeled as extractive keyphrase extraction (EKE) \cite{sun2020joint} whose goal is to identify the salient phrases in a document. These systems are designed to identify the word(s) in a document that forms a phrase and refer to key points in the document. However, there are some differences that render the existing methods for extractive keyphrase extraction inapplicable for our task. First, EKE is conducted on the formal text where the sentences are grammatically correct and no chitchat or repeated information exists in the text. In contrast, in the proposed task we deal with automatically transcribed videos that might be noisy and also it might have redundant information in the forms of chitchat or repeated sentences/words. As such, since EKE systems are not designed to avoid uninformative portions of the input, they will fail on this task. Second, one of the requirements in the proposed task is to ensure the uniqueness of the keyphrases of each paragraph. Existing EKE systems are not designed to observe this requirement, thereby they might extract the same phrase for multiple paragraphs. Third, the existing EKE systems are trained in the scientific or weblogs domain. They are substantially different from the domain of live-stream videos such as Behance videos that generally contain technical content. Moreover, none of the prior works can benefit from the available resources in the general domain (i.e., weblogs) for a domain-specific EKE system. As such, in this work, we propose a novel system that can work on noisy live-stream video transcripts meanwhile benefit from available resources in the general domain. Our proposed system is the first model that is encouraged to ensure the uniqueness of the keyphrases.

\section{Model}

\subsection{Overview}

Our proposed model has the following novelties:

\begin{itemize}
    \item A novel method for extracting keyphrases from the paragraphs of a live-stream video
    \item A novel technique to encourage the uniqueness of the keyphrases across consecutive paragraphs of the transcript
    \item A novel method for identifying the domain-specific phrases using a domain discriminator model
    \item A novel method based on multi-tasking to bridge the phraseness and keyness information from general domain to a specific domain
    \item A novel method based on reinforcement learning to inform the model about the chitchats in the transcript
\end{itemize}

\subsection{Details}

Formally, the input to the system is a paragraph of a transcript, i.e.,  $D=[w_1,w_2,\ldots,w_n]$, consisting of $n$ words. The goal is to select phrases $P$ in $D$ that most clearly represent the main content of the paragraph $D$. A phrase $P$ might be a single word or a span in $D$. All noun, verb, or adjective phrases are eligible to be selected as one of the phrases $P$. In this work, the phrases $P$ are encoded in the label sequence $L^p=[l^p_1,l^p_2,\ldots,l^p_n]$, where $l^p_i \in \{O,B,I\}$ and $B$ indicates the word $w_i$ is the beginning of a phrase and $I$ indicates the word $w_i$ is the continuation of a phrase. To create a system for this task, we propose multiple components. Specifically, the proposed system consists of three major components:

\begin{itemize}
    \item Keyphrase Extractor: The input paragraph $D$ is first encoded to high-dimensional vectors $H=[h_1,h_2,\ldots,h_n]$ using a pre-trained transformer-based language model. Moreover, to preserve the information about the phrases $P'$ extracted from the previous paragraph $D'$, these phrases are also concatenated to the input paragraph $D$. Using the extracted feature vectors for the input words, we predict the label for each word.
    \item Data Augmentation: In addition to the training samples for keyphrase extraction from transcripts, we propose to employ the general domain training samples for keyphrases extraction. However, the domain shift between these resources could impede the training. As such, we devise a component dedicated to bridging the gap between general domain and transcript domain keyphrases. Specifically, this component aims to transfer the keyness and phraseness features learned from the general domain to the domain of transcripts. It is achieved by predicting the bridge phrases. In particular, in addition to the main keyphrase extraction, we suggest training the model to recognize phrases that are representative of their domain (e.g., transcripts or general domain). This task could help the model to get informed of the portions of the input that are specific to the domain of interest, thereby filter out irrelevant knowledge/features related to the other domain. We propose a novel method to obtain a pseudo-label for this task. Specifically, a separate transformer-based language model is trained to discriminate the domains of the paragraphs and the attention scores for each phrase from the discriminator model are employed to construct domain-specific phrases.
    \item Information Filter: In this task, we require the keyphrases to be selected from the informative portion of the input paragraph. That is, chitchats and other non-informative portions should be avoided. In order to encourage the model to observe this requirement, we first create pseudo-binary labels for each sentence of the paragraph where 0 indicates the sentence is chitchat and 1 indicates the sentence is informative. Next, we reward the keyphrase extraction model if the selected keyphrases are from informative sentences.
    % \item Keyphrase Extraction: Finally, using the representations of the words obtained from the input encoder, we attempt to recognize the keyphrases in the input paragraph. In addition to the supervision from the main task, we also reward the model if the selected keyphrases are disjoint from the phrases selected for the previous paragraph.
\end{itemize}

The rest of this section elaborates more on the details of each of these components.

\subsubsection{Keyphrase Extraction}

In this work we use BERT$_{base}$ \cite{devlin2019bert} as the input encoder. It is worth noting that in addition to the input paragraph, we suggest concatenating the keyphrases of the previous paragraph to the input text so that the encoder is aware of the words that it will later be encouraged to avoid selecting them as keyphrases. To this end, to encode the input paragraph $D$ and the keyphrases of the previous paragraph, we construct the sequence $S = [w_1,w_2,\ldots,w_n,SEP,kp_1,kp_2,\ldots,kp_m]$, where $kp_i \in P'$ is the $i-$th keyphrase extracted for the previous paragraph. To find $kp$'s, during training we use the gold keyphrases and during inference, we use the keyphrases predicted by the model. The sequence $S$ is fed into the BERT$_{base}$ model and the representations of the word-pieces of the input paragraph is taken from the final layer of the BERT transformer as the input word representations $H=[h_1,h_2,\ldots,h_n]$. Note that for the words consisting of multiple word pieces we use the average of their word-piece representation to obtain their final vector.

Next, using the vectors $H$, we predict the likelihood of every word to be selected in a keyphrase. Formally, the vector $h_i$ is consumed by a two-layer feed-forward network to estimate the likelihood $P(\cdot|D,w_i.\theta)$:

\begin{equation}
    P(\cdot|D,w_i,\theta) = \sigma(W_2(W_1*h_i+b_1)+b_2,\theta)
\end{equation}
where $\sigma$ is the softmax function, $W_2$ and $W_1$ is the weight matrices, $b_1$ and $b_2$ are the biases, and $P(\cdot|D,w_i,\theta)$ is the label distribution for the word $w_i$ predicted by the model with parameters $\theta$. To train the model for keyphrase extraction, we use the following cross-entropy loss:

\begin{equation}
    \mathcal{L}_{kp} = - \sum_{i=1}^n \log(P(l_i|D,w_i,\theta))
\end{equation}

Also, in order to encourage the model to avoid selecting repeated keyphrases for the consecutive paragraphs, we compute the following reward:

\begin{equation}
        R_{rep}(KP) = - \frac{1}{n} \sum_{i=1}^n \text{REP}(w_i)
\end{equation}
\begin{equation*}
    \text{REP}(w_i) =
    \begin{cases}
        1, & \text{if } w_i \in P' \quad \& \\
         & \text{argmax}(P(\cdot|D,w_i,\theta)) \in \{\text{B,I}\}\\
        0,              & \text{otherwise}
    \end{cases}
\end{equation*}
where $KP$ is the list of predicted keyphrases and $\text{REP}(w_i)$ is a function that returns 1 if the word $w_i$ is predicted to be in a keyphrase and also it appears in the keyphrases of the previous paragraph, i.e., $P'$.

\subsubsection{Data Augmentation}

In this work, we extract keyphrases from live-stream video transcripts. However, there are other manually annotated resources in other domains which might be helpful to improve the performance of the keyphrase extraction model. One of these resources is OpenKP \cite{xiong2019open} that provides human annotation for keyphrase extraction in the domain of web pages. This dataset provides annotation for about seventy thousand web pages. One simple method to employ this resource in the training of the model is to combine this dataset with the annotated samples of the live-stream videos. While this simple technique could help the model to learn more patterns of the keyphrases, one limitation is that the domain shift between web pages and live-stream videos might hinder the model training. Specifically, a model trained on the web page domain might pay more attention to chitchats of the live-stream videos and ignores the informative portion that contains more technical phrases. As such, it is necessary to equip the model with a mechanism to overcome the domain shift. In this work, we suggest that identifying the keyphrases that are representative of their domain is critical information that a model trained on multiple domains should be aware of in order to avoid selecting keyphrases that are suitable for the other auxiliary domain. In other words, we aim to train the model in a multitask setting, in which in addition to the main keyphrase extraction, the model is trained to recognize domain-specific phrases. Therefore, the first step is to create labels for the phrases that are domain-specific in the paragraphs of the live-stream video and the web pages.

\noindent \textbf{Domain-specific Phrase Annotation}: The purpose of domain-specific phrase detection is to recognize the phrases $P' \in D$ that are representative of the domain of $D$. Since there is no labeled data for this task, we resort to an unsupervised method. In the first step, we suggest automatically construct a labeled dataset for domain-specific phrase detection using a pre-trained domain discriminator. In particular, we first combine all training paragraphs $D$ from the live-stream video domain and $D'$ from the web page domain to construct the dataset $\mathcal{D}$. Next, we employ a BERT$_{base}$ model\footnote{Note that it is separate from the main model encoder} to encode the paragraph $\bar{D} \in \mathcal{D}$. The representations $\bar{H}=[\bar{h}_1,\bar{h}_2,\ldots,\bar{h}_n]$ obtained from the last layer of the BERT$_{base}$ model\footnote{We compute the average vector representation for the words with multiple word-pieces} are later max-pooled (mp()) and are consumed by a feed-forward layer to predict the domain of $\bar{D}$:

\begin{equation}
    \begin{split}
        % V & = MAX\_POOL(\bar{h}_1,\bar{h}_2,\ldots,\bar{h}_n) \\
        V & = mp(\bar{h}_1,\bar{h}_2,\ldots,\bar{h}_n) \\
        P(\cdot|\bar{D},\theta') & = \sigma(W_2*(W_1*V+b_1)+b_2)
    \end{split}
\end{equation}
where $\sigma$ is the sigmoid activation function, $W_1$ and $W_2$ are the weight matrices, $b_1$ and $b_2$ are the biases and $P(\cdot|\bar{D},\theta)$ is the probability distribution over the two domain (i.e., live-stream videos and web pages) predicted by the discriminator with parameter $\theta'$. This pre-trained model has an accuracy of 93\% on the test set of $\mathcal{D}$.

Next, we apply the pre-trained domain discriminator on all documents $\bar{D} \in \mathcal{D}$ to obtain the attention scores $A=[a_1,a_2,\ldots,a_n]$ for all words $w_i \in \bar{D}$. Note that these attention scores are obtained from the final layer of BERT$_{base}$ encoder of the discriminator. Using the attention scores $A$, we filter the words of the document $\bar{D} \in \mathcal{D}$. Specifically, in the first step, the attention scores $A$ are sorted descendingly, i.e., $A'=[a'_i | a'_i \in A \quad \& \quad A'_i \geq A'_j \Rightarrow i \leq j]$. We define the function $g(x)$ whose input is the index $i$ of $a'_i$ and its output is the index $j$ of $a_j$ corresponding to entry $a'_i$, i.e. $a_j=a'_i$. Next, the array $W'$ is constructed whose entry $w'_i \in W'$ is the word $w_j \in \bar{D}$ where $j=g(i)$, i.e., the word corresponding to the $i$-th entry of sorted attention score $A'$. Finally, we prune all words of the paragraph $\bar{D}$ that appear after the index $k$ in $W'$:

\begin{equation}
    \hat{\bar{D}} = \{w_i|w_i\in\bar{D}\quad\&\quad\text{Index\_of}(w_i,W') < k\}
\end{equation}
where $\text{Index\_of}(x,Y)$ is a function that returns the index of $x$ in the array $Y$. In order to find the optimal value for $k$, we use the following criteria:

\begin{equation}
    |P(\cdot|\hat{\bar{D}},\theta') - P(\cdot|\bar{D},\theta') | \leq \eta
\end{equation}
where $P(\cdot|\hat{\bar{D}},\theta')$ is the domain distribution predicted by the pre-trained discriminator for the filtered document $\hat{\bar{D}}$, and $\eta$ is a threshold to be selected based on the performance on the development set. The main motivation for this filtering criteria is that only those important words that are necessary to make the same prediction as the original input document should be preserved in the filtered document.

Every word that is remained in the filter document $\hat{\bar{D}}$ is used to create the silver data for domain-specific phrase detection. Concretely, the label vector $L=[l_1,l_2,\ldots,l_n]$ is constructed as the silver labels for domain-specific phrase detection:

\begin{equation}
        l_i =
\begin{cases}
    1, & \text{if } w_i \in \hat{\bar{D}}\\
    0,              & \text{otherwise}
\end{cases}
\end{equation}

Finally, using the silver data, the main keyphrase extraction model is trained to recognize the domain-specific phrases in a multi-task setting. In particular, the vector representation $H=[h_1,h_2,\ldots,h_n]$, obtained from the keyphrase extraction encoder, are fed into a feed-forward layer with sigmoid activation function at the end to predict the domain-specific phrases:

\begin{equation}
    Q(\cdot|D,w_i,\bar{\theta}) = \sigma(W_2(W_1*h_i+b_1)+b_2)
\end{equation}
where $\sigma$ is the sigmoid activation function, $W_1$ and $W_2$ are weight matrices, $b_1$ and $b_2$ are biases and $Q(\cdot|D,w_i,\bar{\theta})$ is the distribution of the labels (i.e., domain-specific or not), predicted by the model with parameter $\bar{\theta}$ for $i-$th word. To train the model using the silver labels $L$, we use the following negative log-likelihood:

\begin{equation}
    \mathcal{L}_{bridge} = -\sum_{i=1}^n \log(Q(l_i|D,w_i,\bar{\theta}))
\end{equation}

\subsubsection{Information Filter}

In a live-stream video, the streamer might diverge from the main content of the video. These portions of the transcripts, which we call chitchats, should be avoided for keyphrase extraction as they are not informative. To this end, we add another component to our model which encourages the model to avoid selecting keyphrases from chitchats. In this work, we propose to do it at the sentence level. That is, keyphrases should not be selected from sentences that are likely to be chitchat. Now, the main question is how can we effectively determine which sentences are chitchat. To answer this question, we resort to an unsupervised method based on the semantic representation of the sentences of the paragraph $D$. Specifically, we hypothesize that the majority of the sentences in the paragraph are on the topic sentences. As such, in order to detect the sentences that are off the topic, we propose to compute the semantic similarity of the sentences with the entire paragraph. Those sentences whose representations are far enough from the paragraph representation are selected as the chitchat sentences. Formally, given the document $D$ encoded by the BERT$_{base}$ model, i.e., the vectors $H$, we take the $[CLS]$ vector representation obtained from the final layer of the BERT transformer as the document level representation, i.e., $h_p$. Next, the representation of the sentence $S_i \in D$ is computed via the max-pooling over its word vector representations:

\begin{equation}
    % h_{S,i} = \text{MAX\_POOL}(\{h_k|w_k\in S_i\})
    h_{S,i} = \text{mp}(\{h_k|w_k\in S_i\})
\end{equation}

Afterward, using the paragraph representation $h_p$ and the sentence representations $h_{S,i}$, we compute a score for every sentence $S_i$:

\begin{equation}
    \alpha_i = \sigma(h_p) \odot \sigma(h_{S,i})
\end{equation}
where $\sigma$ is the softmax operation and $\odot$ is the Hadamard product. Finally, we choose the chitchat sentences based on their computed scores $\alpha_i$:

\begin{equation}
    \text{Is\_Chitchat}(S_i) =
\begin{cases}
    1, & \text{if } \alpha_i \leq \beta\\
    0,              & \text{otherwise}
\end{cases}
\end{equation}
where $\beta$ is a trade-off parameter to be selected based on the performance on the development set. Finally, using the selected chitchat, we define the following reward function:

\begin{equation}
    R_{chitchat}(KP) = - \sum_{i=1} \text{Is\_Chitchat}(\text{Sent}(kp_i))
\end{equation}
where $kp_i \in KP$ is the $i-$th keyphrase selected by the model and $\text{Sent}(x)$ is a function that returns the sentence containing keyphrase $x$.

\subsubsection{Training}

To train the model on the entire task we combined the keyphrase extraction loss $\mathcal{L}_{kp}$, the bridge loss $\mathcal{L}_{bridge}$ and the rewards $R_{rep}(D)$ and $R_{chitchat}(D)$. Since the reward computation is a discrete operation, we resort to Reinforce algorithm to compute the gradients of the rewards. First, the overal reward is computed by $R(KP) = R_{rep}(KP)+\alpha R_{chitchat}(KP)$, where $\alpha$ is a trade-off hyper-parameter. Next, we seek to minimize the negative expected reward $R(KP)$ over the possible choices of $KP$: $\mathcal{L}_{R} = - \mathbb{E}_{\hat{KP} \sim P(\hat{KP}|D)} [R(\hat{KP})]$. The policy gradient is then estimated by: $\nabla \mathcal{L}_{R} = - \mathbb{E}_{\hat{KP} \sim P(\hat{KP}|D)} [(R(\hat{KP}) - b) \nabla \log P(\hat{KP}|D)]$. Using one roll-out sample, we further estimate $\nabla \mathcal{L}_{R}$ via the predicted keyphrases $KP$: $\nabla \mathcal{L}_{R} = -(R(KP) - b) \nabla \log P(\hat{KP}|D)$ where $b$ is the baseline to reduce variance. In this work, we obtain the baseline $b$ via: $b=\frac{1}{|B|}\sum_{i=1}^{|B|} R(KP_i)$, where $|B|$ is the mini-batch size and $KP_i$ is the predicted keyphrases for the $i$-th sample in the mini-batch.

\subsection{Dataset}

To train and evaluate the model we annotate data from the transcripts of the videos streamed on the Behance platform. The recordings are spilled by specialists and creators to share/discuss their inventive ventures. As such, verbal substance from the speakers (in English) is imperative for video understanding. Whereas the recordings have introductory subjects, their substance is impromptu, thus the streamer might cut sentences, examine numerous themes, and utilize casual expressions. The recordings have an normal length of 48 minutes. To get the verbal substance of the gushed recordings, we utilize the Microsoft ASR tool. In add up to, 361 recordings with a add up to length of more than 500 hours are transcribed. A transcript, on normal, contains 7,219 words.

As discussed in the introduction, the long nature of transcripts spurs us to explain keyphrases at two levels. To begin with, at the section level, we characterize a passage in a transcript to have the same part as passages in formal composed archives. Concretely, a passage is characterized
 as a chunk of content that passes on a specific point or thought. A transcript comprises of numerous disjoint passages. Since the ASR content does not give section data, we clarify the collected transcripts with sections. Next, for each section of the transcript, the critical keyphrases are chosen. To this conclusion, a keyphrase for a passage ought to have these features: (a) Concisely summarize the most idea within the section; (b) Be related to the most subject of the video; (c) Expressly show up within the passage; (d) Does not appear within the past or following passages; (e) Shape a legitimate English noun/verb state. The passages that are totally off-topic don't have any keyphrase. Next, at the chapter level, we offer keyphrases for chapters within the transcripts. A chapter comprises of numerous sections to speak to a single point. For instance, in a photo altering video, the talk on how to alter the foundation can shape a chapter. A keyphrase of a chapter should observe the following criteria: (a) Concisely summarize the most subjects within the chapter; (b) May not unequivocally show up within the chapter; (c) Does not cover with the section keyphrases or other chapter level keyphrases; (d) Shape a appropriate English noun/verb state. Note that passages and chapters might have different keyphrases that are sorted based on their significance.

 To explain information for each level, we enlist 10 annotators from the {\tt upwork.com} platform which is a site for enlisting freelancers
 with distinctive mastery. Since the collected recordings are related to photo altering program, e.g., Photoshop, we require the annotators to have encounter both in information explanation and in utilizing major photo altering apparatuses. We prepare the annotators for KP at each level. To anticipate chapter-level keyphrases to be one-sided toward paragraph-level keyphrases, we part annotator pool for section and chapter level explanation (five for each). The transcripts are disseminated equally to the five annotators at each level for comment. As such, a transcript is explained completely by a passage annotator and a chapter annotator.

\begin{figure*}
     \centering
     \resizebox{.95\textwidth}{!}{
     \includegraphics{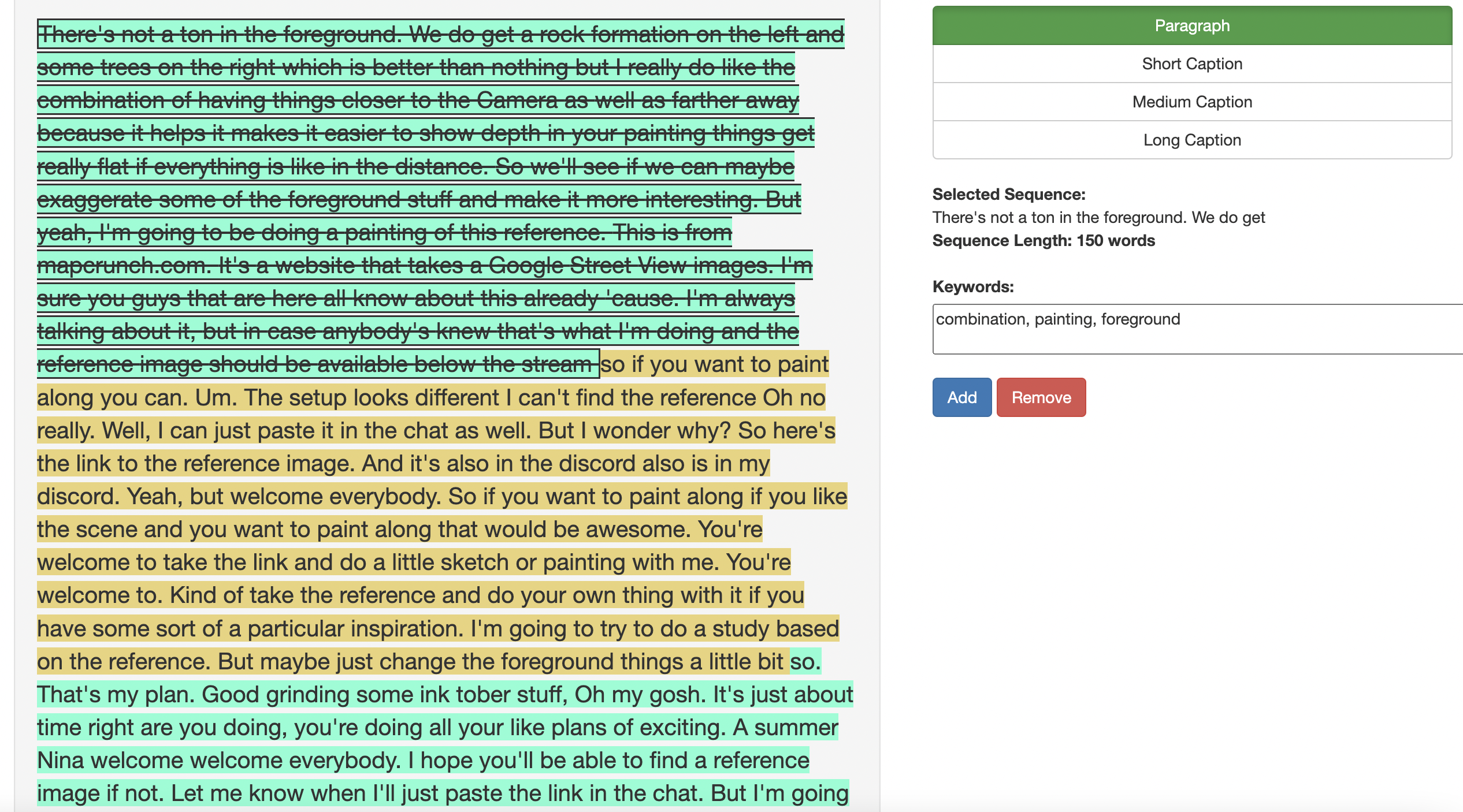}
     }
     \caption{Screenshot of the annotation tool}
     \label{fig:tool}
 \end{figure*}

 Figure \ref{fig:tool} shows an example of the annotation tool. In this tool the annotator can select/correct the boundaries of the paragraphs in each video transcripts. Each paragraph are shown in different colors. For each paragraph, the annotator selects a group of the keyphrases fron the paragraph. In the example shown in Figure \ref{fig:tool}, the keyphrases ``\textit{combination, painting, foreground }" are provided. For more examples of the annotation tool, see appendices.

\subsubsection{Results}

\begin{table}[]
    \centering
    \resizebox{.45\textwidth}{!}{
    \begin{tabular}{l|c|c|c}
       Model & F1@1  & F1@3 & F1@5 \\ \hline
       JointKPE (BERT)  & 14.44  & 18.91 & 24.19 \\
       JointKPE (RoBERTa) & 16.00 & 22.07 & 25.08 \\
       JointKPE (SpanBERT) & 16.08 & 24.96 & 27.63 \\ \hline
       Ours & 28.50 & 36.43 & 33.83
    \end{tabular}
    }
    \caption{Performance of the models}
    \label{tab:results}
\end{table}

We evaluate the performance of the keyphrase extraction model on the dataset of live-stream video transcripts. In our experiments, we compare our model with the previous state-of-the-art keyphrase extraction model JointKPE \cite{sun2020joint}. We compare the models using the F1 score at the first 1, 3, and 5 keyphrases extracted by the model\footnote{Keyphrases are sorted based on the likelihood of their first word, i.e., $P(B|D,w_i,\theta)$}. Table \ref{tab:results} show the results of this comparison. As shown in this table, the proposed model significantly outperforms the baselines in all metrics. This superior performance demonstrates the effectiveness of the proposed model.

% Use \bibliography{yourbibfile} instead or the References section will not appear in your paper
\bibliography{aaai22}

 \clearpage
 \clearpage
\newpage

\section{Appendices}

\begin{figure*}
    \centering
    \resizebox{1.0\textwidth}{!}{
    \includegraphics{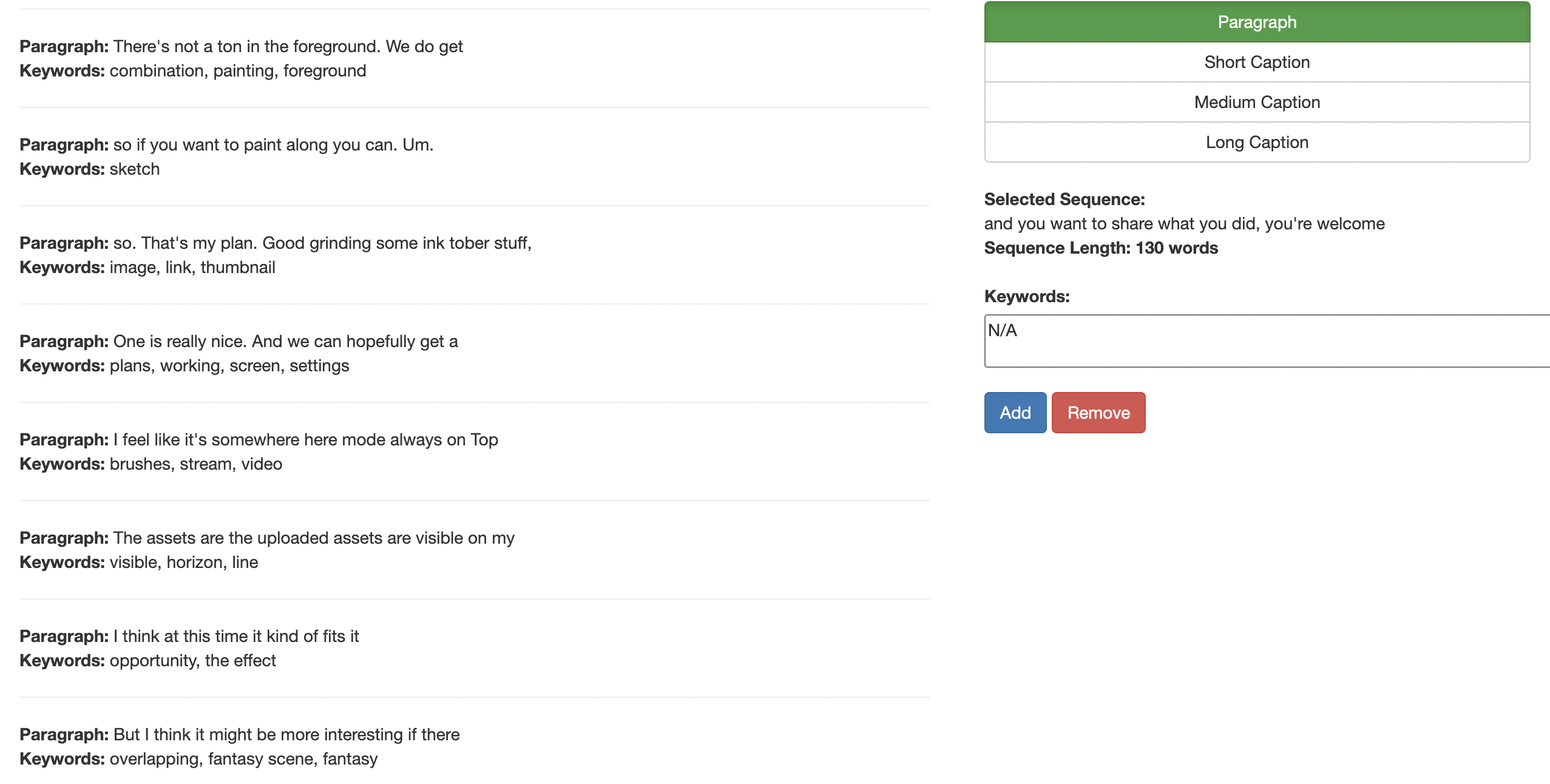}
    }
    \caption{Annotation tool for providing the keyphrases of the entire transcript}
    \label{fig:tool2}
\end{figure*}

\section{Annotation Tool}

An example of the annotation tool for providing the summeray of the selected keyphrases of the entire transcript is shown in Figure \ref{fig:tool2}. In particular, for each annotated paragraph in the given transcript, the annotators see a list of provided keyphrases. Note that in the case that the paragraph is entirely chitchat or off-topic, the annotator provide ``\textit{N/A}" keyphrase.

\section{Case Study}

\begin{figure*}
    \centering
    \resizebox{.55\textwidth}{!}{
    \includegraphics{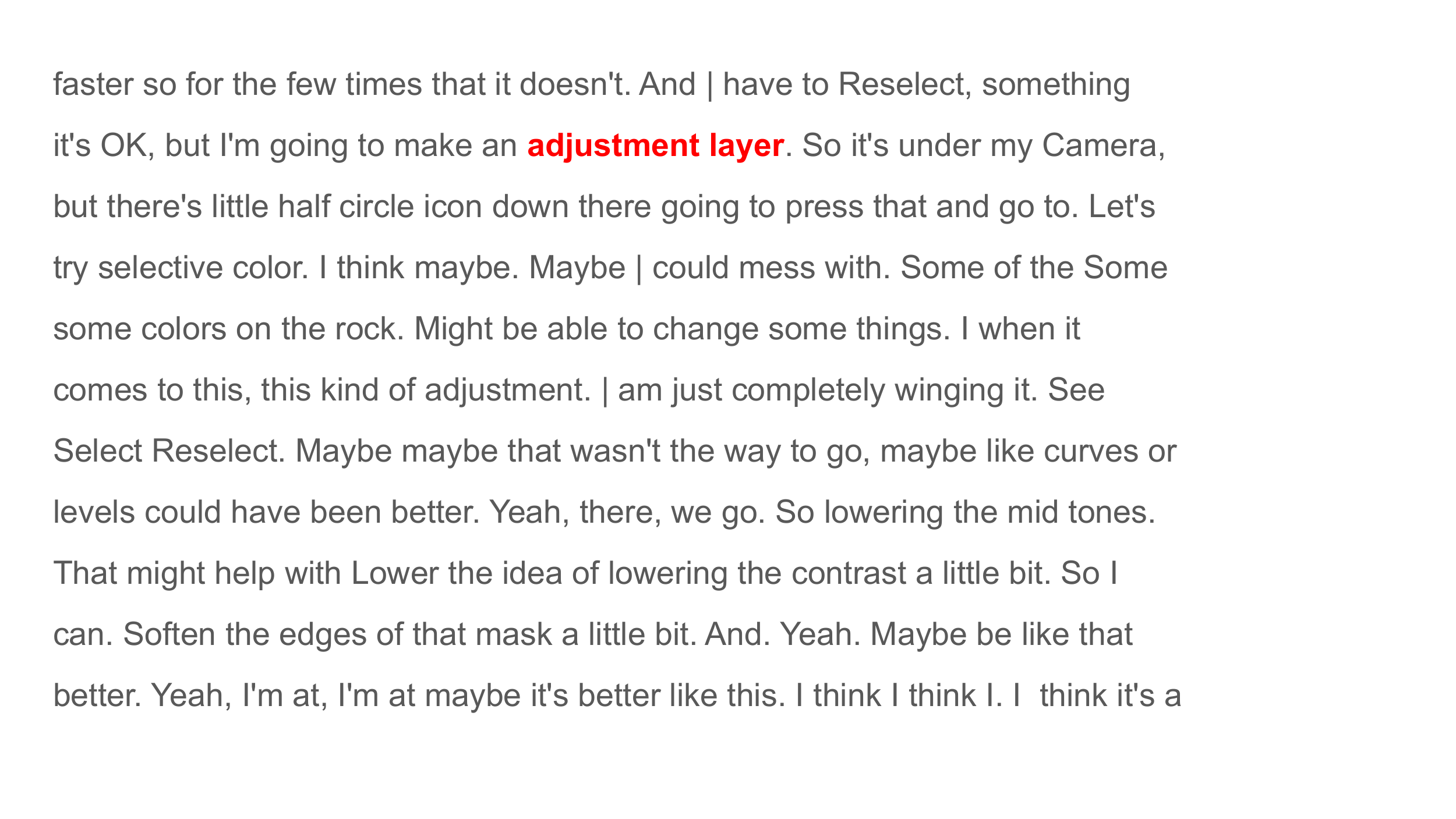}
    }
    \caption{Case Study - The keyphrase is shown in red bold-face}
    \label{fig:case1}
\end{figure*}

\begin{figure*}
    \centering
    \resizebox{.55\textwidth}{!}{
    \includegraphics{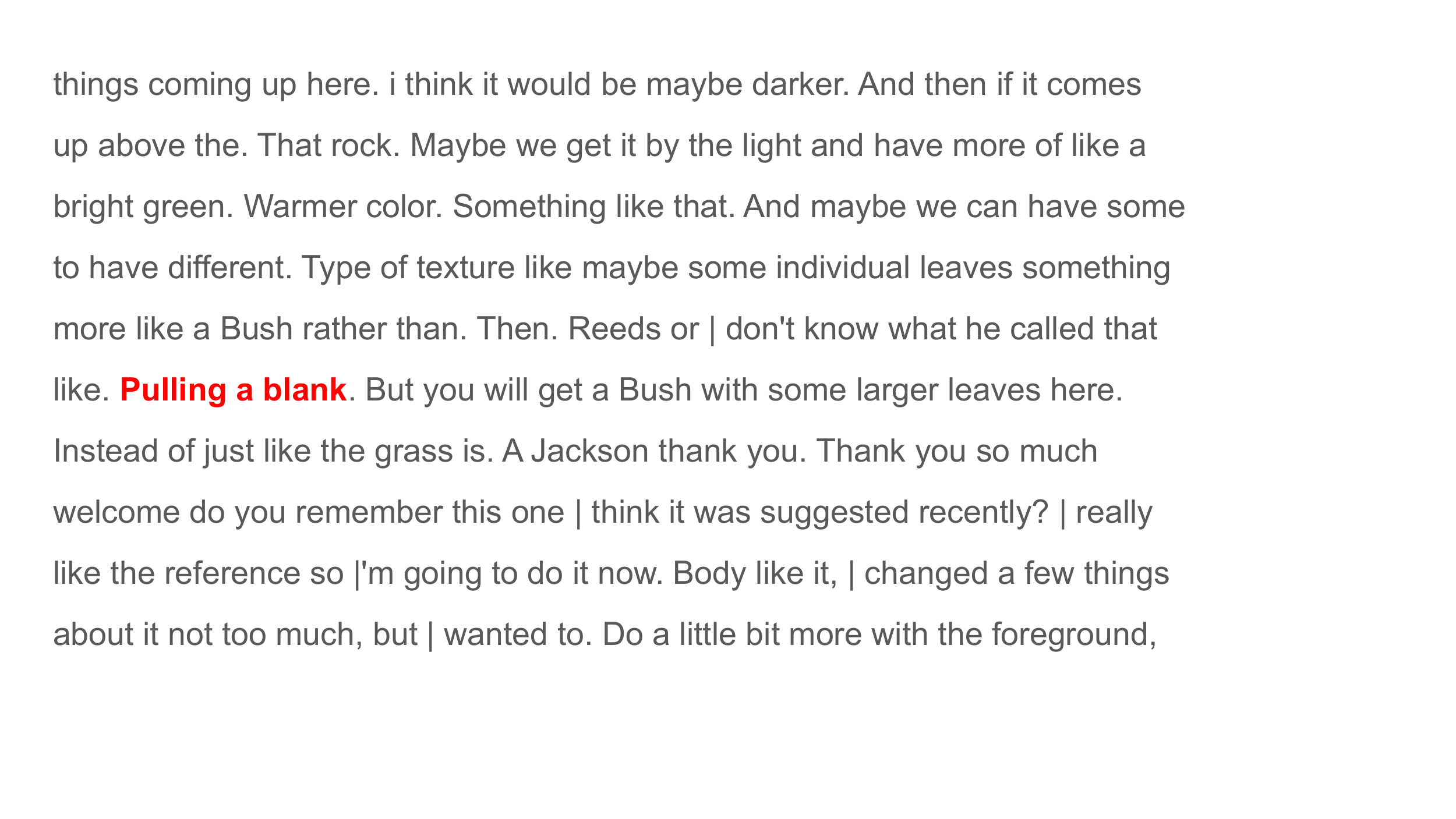}
    }
    \caption{Case Study - The keyphrase is shown in red bold-face}
    \label{fig:case2}
\end{figure*}

\begin{figure*}
    \centering
    \resizebox{.55\textwidth}{!}{
    \includegraphics{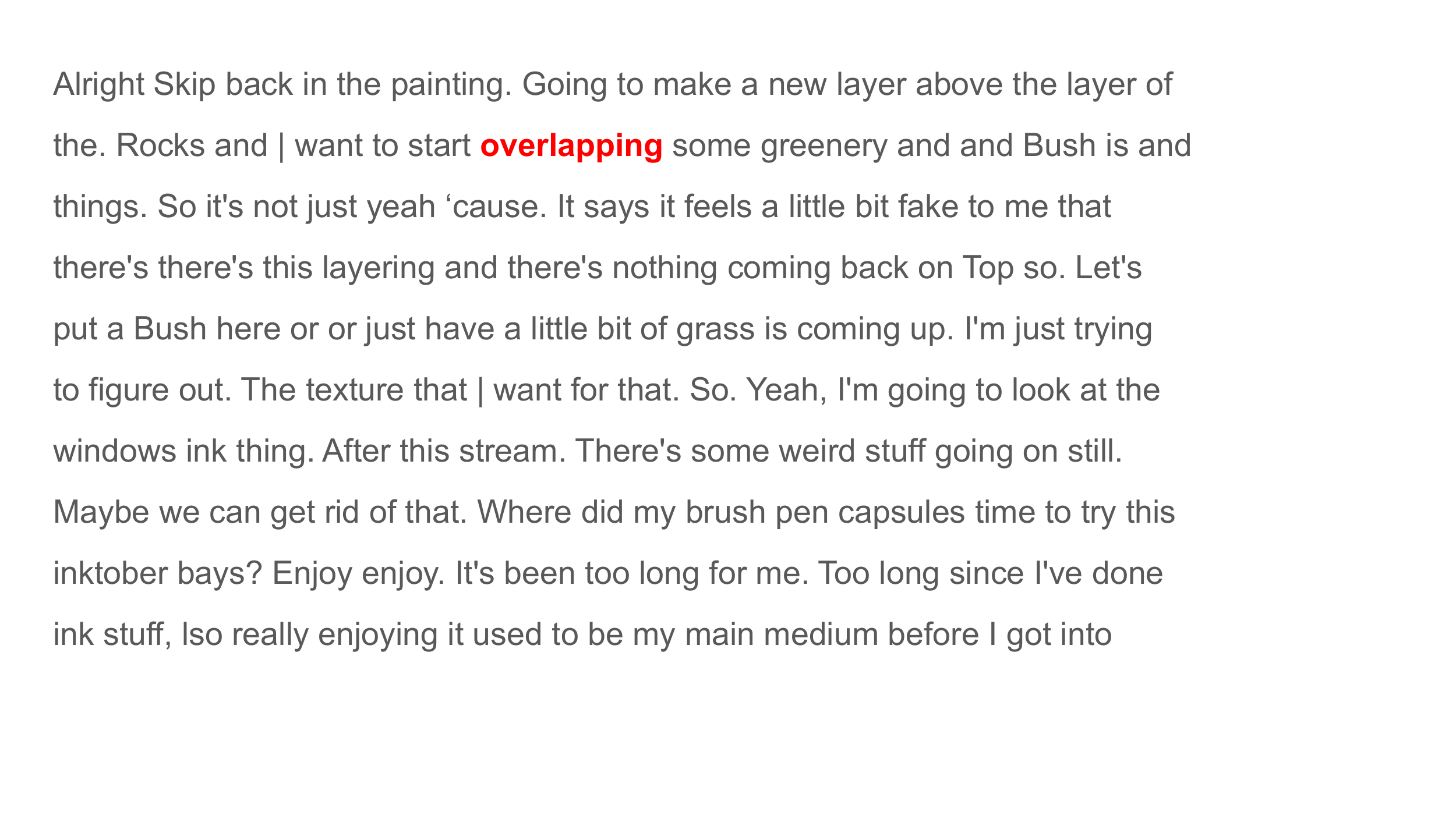}
    }
    \caption{Case Study - The keyphrase is shown in red bold-face}
    \label{fig:case3}
\end{figure*}

To shed more light in the performance of the prsented model, in this section we provide some examples of the extracted keyphrases by the model. Figures \ref{fig:case1}, \ref{fig:case2}, and \ref{fig:case3} show the examples. In these examples, the predicted keyphrase are shown in red bold-face.

\end{document}